\documentclass[10pt,twocolumn,letterpaper]{article}

\usepackage{cvpr}              
\usepackage[accsupp]{axessibility}

\usepackage{adjustbox}
\usepackage{multirow}
\usepackage{tabularx}
\usepackage{makecell}
\usepackage[accsupp]{axessibility} 

\definecolor{cvprblue}{rgb}{0.21,0.49,0.74}
\usepackage[pagebackref,breaklinks,colorlinks,citecolor=cvprblue]{hyperref}

\def\paperID{*****} 
\def\confName{CVPR}
\def\confYear{2026}

\title{The Fourth Challenge on Image Super-Resolution ($\times$4) at NTIRE 2026: Benchmark Results and Method Overview}

\author{
Zheng Chen$^{\dagger}$ \and
Kai Liu$^{\dagger}$ \and
Jingkai Wang$^{\dagger}$ \and
Xianglong Yan$^{\dagger}$ \and
Jianze Li$^{\dagger}$ \and
Ziqing Zhang$^{\dagger}$ \and
Jue Gong$^{\dagger}$ \and
Jiatong Li$^{\dagger}$ \and
Lei Sun$^{\dagger}$ \and
Xiaoyang Liu$^{\dagger}$ \and
Radu Timofte$^{\dagger}$ \and
Yulun Zhang$^{\dagger*}$ \and
Jihye Park \and
Yoonjin Im \and
Hyungju Chun \and
Hyunhee Park \and
MinKyu Park \and
Zheng Xie \and
Xiangyu Kong \and
Weijun Yuan \and
Zhan Li \and
Qiurong Song \and
Luen Zhu \and
Fengkai Zhang \and
Xinzhe Zhu \and
Junyang Chen \and
Congyu Wang \and
Yixin Yang \and
Zhaorun Zhou \and
Jiangxin Dong \and
Jinshan Pan \and
Shengwei Wang \and
Jiajie Ou \and
Baiang Li \and
Sizhuo Ma \and
Qiang Gao \and
Jusheng Zhang \and
Jian Wang \and
Keze Wang \and
Yijiao Liu \and
Yingsi Chen \and
Hui Li \and
Yu Wang \and
Congchao Zhu \and
Saeed Ahmad \and
Ik Hyun Lee \and
Jun Young Park \and
Ji Hwan Yoon \and
Kainan Yan \and
Zian Wang \and
Weibo Wang \and
Shihao Zou \and
Chao Dong \and
Wei Zhou \and
Linfeng Li \and
Jaeseong Lee \and
Jaeho Chae \and
Jinwoo Kim \and
Seonjoo Kim \and
Yucong Hong \and
Zhenming Yan \and
Junye Chen \and
Ruize Han \and
Song Wang \and
Yuxuan Jiang \and
Chengxi Zeng \and
Tianhao Peng \and
Fan Zhang \and
David Bull \and
Tongyao Mu \and
Qiong Cao \and
Yifan Wang \and
Youwei Pan \and
Leilei Cao \and
Xiaoping Peng \and
Wei Deng \and
Yifei Chen \and
Wenbo Xiong \and
Xian Hu \and
Yuxin Zhang \and
Xiaoyun Cheng\and
Yang Ji \and
Zonghao Chen \and
Zhihao Xue \and
Junqin Hu \and
Nihal Kumar \and
Snehal Singh Tomar \and
Klaus Mueller \and
Surya Vashisth \and
Prateek Shaily \and
Jayant Kumar \and
Hardik Sharma \and
Ashish Negi \and
Sachin Chaudhary \and
Akshay Dudhane \and
Praful Hambarde \and
Amit Shukla \and
Shijun Shi \and
Jiangning Zhang \and
Yong Liu \and
Kai Hu \and
Jing Xu \and
Xianfang Zeng \and
Amitesh M \and
Hariharan S \and
Chia-Ming Lee \and
Yu-Fan Lin \and
Chih-Chung Hsu \and
Nishalini K \and
Sreenath K A \and
Bilel Benjdira \and
Anas M. Ali \and
Wadii Boulila \and
Shuling Zheng \and
Zhiheng Fu \and
Feng Zhang \and
Zhanglu Chen \and
Boyang Yao \and
Nikhil Pathak \and
Aagam Jain \and
Milan Kumar \and
Kishor Upla \and
Vivek Chavda \and
Sarang N S \and
Raghavendra Ramachandra \and
Zhipeng Zhang \and
Qi Wang \and
Shiyu Wang \and
Jiachen Tu \and
Guoyi Xu \and
Yaoxin Jiang \and
Jiajia Liu \and
Yaokun Shi \and
Yuqi Li \and
Chuanguang Yang \and
Weilun Feng \and
Zhuzhi Hong \and
Hao Wu \and
Junming Liu \and
Yingli Tian \and
Amish Bhushan Kulkarni \and
Tejas R R Shet \and
Saakshi M Vernekar \and
Nikhil Akalwadi \and
Kaushik Mallibhat \and
Ramesh Ashok Tabib \and
Uma Mudenagudi \and
Yuwen Pan \and
Tianrun Chen \and
Deyi Ji \and
Qi Zhu \and
Lanyun Zhu \and
Heyan Zhangyi
}

\begin{document}

\maketitle

\begingroup
\renewcommand\thefootnote{}
\footnotetext{$^{\dagger}$Zheng Chen, Kai Liu, Jingkai Wang, Xianglong Yan, Jianze Li, Ziqing Zhang, Jue Gong, Jiatong Li, Lei Sun, Xiaoyang Liu, Radu Timofte and Yulun Zhang are the challenge organizers, while the other authors participated in the challenge. $^{*}$Corresponding author: Yulun Zhang. Section~\textcolor{red}{B} in the supplementary materials contains the authors' teams and affiliations. NTIRE 2026 webpage: \url{https://cvlai.net/ntire/2026}. Code: \url{https://github.com/zhengchen1999/NTIRE2026_ImageSR_x4}.}
\endgroup

\begin{abstract}
This paper presents the NTIRE 2026 image super-resolution ($\times$4) challenge, one of the associated competitions of the NTIRE 2026 Workshop at CVPR 2026. The challenge aims to reconstruct high-resolution (HR) images from low-resolution (LR) inputs generated through bicubic downsampling with a $\times$4 scaling factor. The objective is to develop effective super-resolution solutions and analyze recent advances in the field. To reflect the evolving objectives of image super-resolution, the challenge includes two tracks: \textbf{(1) a restoration track}, which emphasizes pixel-wise fidelity and ranks submissions based on PSNR; and \textbf{(2) a perceptual track}, which focuses on visual realism and evaluates results using a perceptual score. A total of 194 participants registered for the challenge, with 31 teams submitting valid entries. This report summarizes the challenge design, datasets, evaluation protocol, main results, and methods of participating teams. The challenge provides a unified benchmark and offers insights into current progress and future directions in image super-resolution.
\end{abstract}

\vspace{-4.mm}
\section{Introduction}

Single image super-resolution (SR) aims to reconstruct a high-resolution (HR) image from a low-resolution (LR) input. The LR image is obtained through an information-losing degradation process. SR is a fundamental problem in computer vision. It supports many applications, such as surveillance, medical imaging, and remote sensing~\cite{zou2012very,shi2013cardiac}. Among different settings, the classical SR task~\cite{timofte2017ntire} is the most widely used benchmark for image SR.

In this setting, LR images are generated by a predefined downsampling process. Bicubic interpolation is the most common choice. This process removes high-frequency details. It makes SR an ill-posed problem. The goal is to recover missing details using learned priors~\cite{zhang2018image}. The classical setting is simple and clear. It allows fair comparison between methods. Models trained in this setting often generalize well to more complex degradations~\cite{khan2022ntire}.

Early SR methods rely on interpolation and reconstruction techniques~\cite{capel2003computer,zibetti2007robust,zhang2012single}. These methods produce smooth results. They fail to recover fine textures. Deep learning changes this trend. Convolutional neural networks have become the dominant solution~\cite{dong2014learning,zhang2018residual,zhang2018image,dai2019secondSAN,mei2020imageCSNLN}. Starting from SRCNN~\cite{dong2014learning}, later works introduce deeper networks, residual connections, and attention mechanisms~\cite{he2016deep,zhang2019rnan,niu2020singleHAN,mei2020imageCSNLN,chen2021attention}. Transformer-based models further improve performance. They capture long-range dependencies using self-attention~\cite{liu2021swin,chen2023activating,chen2024recursive}. Recent sequence models, such as Mamba~\cite{gu2023mamba,guo2024mambair,mambaIRv2}, provide better efficiency. Most of these methods focus on pixel-wise reconstruction.

Recently, the focus shifts to perceptual quality. GAN-based methods generate more realistic textures~\cite{goodfellow2014generative,wang2021real,zhang2017image}. Diffusion models provide another powerful solution~\cite{rombach2022high,ramesh2022hierarchical,blattmann2023stable}. They generate images through a denoising process. They better model complex distributions. Recent works further explore efficient one-step diffusion-based SR~\cite{wang2024sinsr,wu2024one,dong2025tsd,li2025one,li2025distillation}, which aims to reduce sampling cost. These methods improve visual realism but may reduce fidelity. As a result, SR now involves a trade-off between distortion and perception.

Besides model design, many works improve SR through training and inference strategies. Pre-trained models are widely used~\cite{rombach2022high,flux2}. Large-scale datasets also play an important role. Advanced loss functions are introduced to enhance details. Inference-time techniques, such as self-ensemble~\cite{timofte2016seven}, further boost performance. These strategies are widely adopted in recent competitions.

We organize the NTIRE 2026 Challenge on image super-resolution ($\times4$), following previous editions~\cite{ntire2023srx4,ntire2024srx4,ntire2025srx4}. The challenge follows the classical bicubic setting. It aims to recover HR images from LR inputs. We use the DIV2K test dataset~\cite{timofte2017ntire} for evaluation. The goal is to benchmark state-of-the-art methods and analyze recent progress.

The challenge includes two tracks. The first track focuses on restoration quality. It uses PSNR as the main metric. The second track focuses on perceptual quality. It uses multiple image quality metrics for evaluation. This design reflects the dual objectives of SR. It encourages methods that balance fidelity and perceptual quality.

This challenge is one of the challenges associated with the NTIRE 2026 Workshop~\footnote{\url{https://www.cvlai.net/ntire/2026/}} on:
deepfake detection~\cite{ntire26deepfake}, 
high-resolution depth~\cite{ntire26hrdepth},
multi-exposure image fusion~\cite{ntire26raim_fusion}, 
AI flash portrait~\cite{ntire26raim_portrait}, 
professional image quality assessment~\cite{ntire26raim_piqa},
light field super-resolution~\cite{ntire26lightsr},
3D content super-resolution~\cite{ntire263dsr},
bitstream-corrupted video restoration~\cite{ntire26videores},
X-AIGC quality assessment~\cite{ntire26XAIGCqa},
shadow removal~\cite{ntire26shadow},
ambient lighting normalization~\cite{ntire26lightnorm},
controllable Bokeh rendering~\cite{ntire26bokeh},
rip current detection and segmentation~\cite{ntire26ripdetseg},
low light image enhancement~\cite{ntire26llie},
high FPS video frame interpolation~\cite{ntire26highfps},
Night-time dehazing~\cite{ntire26nthaze,ntire26nthaze_rep},
learned ISP with unpaired data~\cite{ntire26isp},
short-form UGC video restoration~\cite{ntire26ugcvideo},
raindrop removal for dual-focused images~\cite{ntire26dual_focus},
image super-resolution (x4)~\cite{ntire26srx4},
photography retouching transfer~\cite{ntire26retouching},
mobile real-word super-resolution~\cite{ntire26rwsr},
remote sensing infrared super-resolution~\cite{ntire26rsirsr},
AI-Generated image detection~\cite{ntire26aigendet},
cross-domain few-shot object detection~\cite{ntire26cdfsod},
financial receipt restoration and reasoning~\cite{ntire26finrec},
real-world face restoration~\cite{ntire26faceres},
reflection removal~\cite{ntire26reflection},
anomaly detection of face enhancement~\cite{ntire26anomalydet},
video saliency prediction~\cite{ntire26videosal},
efficient super-resolution~\cite{ntire26effsr},
3d restoration and reconstruction in adverse conditions~\cite{ntire26realx3d},
image denoising~\cite{ntire26denoising},
blind computational aberration correction~\cite{ntire26aberration},
event-based image deblurring~\cite{ntire26eventblurr},
efficient burst HDR and restoration~\cite{ntire26bursthdr},
low-light enhancement: `twilight cowboy'~\cite{ntire26twilight},
and efficient low light image enhancement~\cite{ntire26effllie}.

\begin{table*}[t]
    \centering
    \setlength{\tabcolsep}{1.3mm}
    \begin{adjustbox}{width=\linewidth}
    \begin{tabular}{l|cc|ccccccccc}
        \toprule
        \multirow{2}{*}{Team Name} & {Rank} & {Rank} & {PSNR} & \multirow{2}{*}{SSIM } & \multirow{2}{*}{LPIPS } & \multirow{2}{*}{DISTS } & \multirow{2}{*}{NIQE } & \multirow{2}{*}{ManIQA } & \multirow{2}{*}{MUSIQ } & \multirow{2}{*}{CLIP-IQA } & {Perceptual Score} \\
        & \textbf{Track 1} & \textbf{Track 2} & \textbf{Track 1} & & & & & & & & \textbf{Track 2}  \\
        \midrule
        SamsungAICamera   &  1 &  1 & 33.73 & 0.9115 & 0.1493 & 0.0648 & 3.6155 & 0.6320 & 76.2912 & 0.9660 & 4.7853 \\
        I2WM\&JNU         &  2 & 10 & 33.45 & 0.9124 & 0.1688 & 0.0931 & 4.4132 & 0.3565 & 62.0699 & 0.4930 & 3.7671 \\
        VEPG              & 25 &  2 & 25.61 & 0.7305 & 0.1718 & 0.0839 & 2.9199 & 0.6495 & 71.6392 & 0.9485 & 4.7666 \\
        SR-Strugglers     &  3 & 12 & 31.98 & 0.8788 & 0.2106 & 0.1175 & 5.1431 & 0.3707 & 61.1988 & 0.5197 & 3.6600 \\
        HONORAICamera     & 24 &  3 & 25.70 & 0.7361 & 0.2020 & 0.1053 & 3.2589 & 0.5286 & 69.6788 & 0.8865 & 4.4787 \\
        IK-Lab            &  4 & 23 & 31.18 & 0.8653 & 0.2306 & 0.1246 & 5.4341 & 0.3578 & 59.3553 & 0.4980 & 3.5507 \\
        RandomSeed42      & 26 &  4 & 24.58 & 0.7003 & 0.2314 & 0.1410 & 3.9686 & 0.6092 & 71.1567 & 0.8401 & 4.3917 \\
        FengFans          &  5 & 18 & 31.18 & 0.8654 & 0.2277 & 0.1212 & 5.3560 & 0.3601 & 59.7416 & 0.5027 & 3.5757 \\
        CIPLAB            & 31 &  5 & 21.77 & 0.5673 & 0.2035 & 0.1258 & 3.0903 & 0.4665 & 72.7605 & 0.7382 & 4.2940 \\
        SUAT              &  6 & 15 & 31.15 & 0.8654 & 0.2260 & 0.1230 & 5.3540 & 0.3631 & 59.9209 & 0.5022 & 3.5801 \\
        BVISR             & 28 &  6 & 23.22 & 0.6460 & 0.2833 & 0.1464 & 3.1647 & 0.5382 & 71.5558 & 0.7789 & 4.2865 \\
        Earth4D           &  7 & 13 & 31.12 & 0.8650 & 0.2241 & 0.1226 & 5.3126 & 0.3670 & 60.2414 & 0.5047 & 3.5961 \\
        TranssionAI       & 27 &  7 & 23.42 & 0.6502 & 0.2723 & 0.1266 & 2.7078 & 0.5057 & 70.4828 & 0.7414 & 4.2822 \\
        AxeraAI           &  8 & 16 & 31.10 & 0.8645 & 0.2261 & 0.1231 & 5.3600 & 0.3632 & 59.8967 & 0.5002 & 3.5772 \\
        MIPLUSCV          & 29 &  8 & 23.08 & 0.6668 & 0.2457 & 0.1211 & 3.1729 & 0.5122 & 70.1923 & 0.7364 & 4.2664 \\
        cialloworld       &  9 & 19 & 31.10 & 0.8645 & 0.2278 & 0.1232 & 5.3572 & 0.3600 & 59.6347 & 0.4985 & 3.5681 \\
        AIT               & 30 &  9 & 22.78 & 0.6129 & 0.3871 & 0.1800 & 3.0319 & 0.5131 & 70.1016 & 0.7456 & 4.0894 \\
        VAI-GM            & 10 & 20 & 31.03 & 0.8636 & 0.2275 & 0.1231 & 5.3591 & 0.3581 & 59.5120 & 0.4992 & 3.5659 \\
        scrlb             & 11 & 22 & 31.02 & 0.8632 & 0.2295 & 0.1250 & 5.3970 & 0.3581 & 59.4856 & 0.4962 & 3.5550 \\
        APRIL-AIGC        & 23 & 11 & 28.24 & 0.7931 & 0.1906 & 0.0998 & 4.0970 & 0.3041 & 57.3545 & 0.5048 & 3.6823 \\
        AH-SNU            & 12 & 21 & 31.01 & 0.8627 & 0.2301 & 0.1253 & 5.3901 & 0.3607 & 59.6212 & 0.5005 & 3.5630 \\
        ACVLAB            & 13 & 14 & 30.82 & 0.8635 & 0.2302 & 0.1210 & 5.2777 & 0.3642 & 59.9242 & 0.5071 & 3.5916 \\
        GLASSv2           & 14 & 28 & 30.53 & 0.8533 & 0.2502 & 0.1316 & 5.6951 & 0.3287 & 55.5208 & 0.4628 & 3.3955 \\
        PSU               & 15 & 24 & 30.51 & 0.8545 & 0.2456 & 0.1297 & 5.4897 & 0.3542 & 58.8557 & 0.4962 & 3.5147 \\
        JNU620            & 16 & 26 & 30.47 & 0.8515 & 0.2465 & 0.1346 & 5.6613 & 0.3423 & 57.4555 & 0.4826 & 3.4523 \\
        Anant\_SVNIT      & 19 & 17 & 29.75 & 0.8636 & 0.2278 & 0.1236 & 5.3596 & 0.3588 & 59.6743 & 0.5089 & 3.5771 \\
        AIMLAB            & 17 & 25 & 30.44 & 0.8514 & 0.2294 & 0.1121 & 5.1282 & 0.3330 & 57.4286 & 0.4451 & 3.4982 \\
        NTR               & 18 & 27 & 30.36 & 0.8488 & 0.2513 & 0.1346 & 5.6802 & 0.3432 & 57.9384 & 0.4788 & 3.4474 \\
        NoReject          & 20 & 29 & 28.86 & 0.8090 & 0.2795 & 0.1590 & 5.9720 & 0.3292 & 54.5854 & 0.4155 & 3.2548 \\
        KLETech-CEVI      & 21 & 31 & 28.68 & 0.8003 & 0.3456 & 0.1748 & 7.0526 & 0.2654 & 38.0155 & 0.3897 & 2.8096 \\
        SFVision          & 22 & 30 & 28.33 & 0.7910 & 0.3909 & 0.1859 & 7.3839 & 0.2884 & 33.8442 & 0.5278 & 2.8395 \\
        \bottomrule
    \end{tabular}
    \end{adjustbox}
    \caption{\textbf{Results of NTIRE 2026 Image Super-Resolution ($\times$4) Challenge.} PSNR and seven perceptual metric scores are evaluated on the DIV2K test set (100 images). In Track 1 (restoration quality), rankings are determined by PSNR values. In Track 2 (perceptual quality), rankings are based on the perceptual score, computed as a weighted combination of seven perceptual metrics.
    The overall ranking prioritizes the better performance between the two tracks, with ties resolved by the average ranking across both tracks.
    The Team descriptions in the main paper Sec.~\ref{sec:teams} and supplementary material Sec.~\textcolor{red}{A} are ordered according to this table.}
    \label{tab:main_results}
\end{table*}

\section{NTIRE 2026 Image Super-Resolution ($\times$4)}
The NTIRE 2026 Image Super-Resolution ($\times$4) Challenge is one of the associated challenges of NTIRE 2026, pursuing two main goals. First, it seeks to deliver a broad and up-to-date overview of recent progress and emerging directions in image super-resolution (SR). Second, it provides a venue where academic researchers and industry practitioners can come together and explore opportunities for collaboration. The sections below describe the specific details.

\vspace{2.mm}
\subsection{Dataset}
\vspace{1.mm}
Two official datasets are provided for this challenge: DIV2K~\cite{timofte2017ntire} and LSDIR~\cite{li2023lsdir}. The use of additional supplementary data for training is also permitted. LR-HR image pairs are generated from HQ images through bicubic interpolation with a $\times$4 downsampling factor.

\noindent\textbf{DIV2K.}
The DIV2K dataset consists of 1,000 2K resolution images, split into three subsets of 800, 100, and 100 images for training, validation, and testing, respectively. To maintain fairness, the high-resolution (HR) images corresponding to the DIV2K validation set are withheld from participants until the testing phase begins. The HR test images are likewise kept hidden for the duration of the entire challenge.

\noindent\textbf{LSDIR.}
The LSDIR dataset contains 86,991 high-quality images sourced from the Flickr platform, partitioned into three subsets of 84,991, 1,000, and 1,000 images for training, validation, and testing, respectively.

\vspace{2.mm}
\subsection{Track and Competition}
\vspace{2.mm}
\label{sec:track}
This year, the competition features two tracks: the Restoration track and the Perceptual track.

\noindent\textbf{Restoration Track.} Following the same protocol as last year's challenge~\cite{ntire2025srx4}, all teams are ranked according to the PSNR computed between their enhanced HR images and the ground-truth HR images from the DIV2K test dataset.

\noindent\textbf{Perceptual Track.} Following last year's challenge~\cite{ntire2025srx4}, six widely used IQA metrics are employed to comprehensively assess the restored results. These metrics are LPIPS, DISTS, CLIP-IQA, MANIQA, MUSIQ, and NIQE. The final ranking of teams is determined by the perceptual score:
\begin{equation}
\begin{aligned}
        \text{Score} = \left(1 - \text{LPIPS}\right) + \left(1 - \text{DISTS}\right) + \text{CLIP-IQA} \\
        + \text{MANIQA} + \frac{\text{MUSIQ}}{100} + \max\left(0, \frac{10 - \text{NIQE}}{10}\right).
\end{aligned}
\end{equation}

\noindent\textbf{Challenge Phases.}
\textit{(1) Development and Validation Phase:} During this phase, participants are given access to two datasets: (a) 800 LR/HR training image pairs and 100 LR validation images from the DIV2K dataset, and (b) 84,991 LR/HR training image pairs and 1,000 LR validation images from the LSDIR dataset. The use of additional external data for training is also permitted. Participants may upload their restored HR images to the Codabench server, where they are evaluated against eight performance metrics and receive immediate feedback.

\textit{(2) Testing Phase:} In the final phase, participants are provided with 100 LR test images without the corresponding HR ground truth images. They are required to submit their SR outputs to the Codabench server, along with their code and a detailed report sent to the organizers via email. Once the challenge concludes, the organizers will validate the submitted code and communicate the final results.

\noindent\textbf{Evaluation Protocol.} Eight standard metrics are adopted for evaluation: PSNR, SSIM, LPIPS, DISTS, NIQE, ManIQA, MUSIQ, and CLIP-IQA. A 4-pixel border is excluded from each image during evaluation, and all calculations are carried out on the Y channel of the YCbCr color space. The evaluation results are primarily determined by the submissions made to the Codabench server, while the submitted code is used for reproduction and verification, with minor precision discrepancies being deemed acceptable. An evaluation script for these metrics is available at \url{https://github.com/zhengchen1999/NTIRE2026_ImageSR_x4}, which also contains the source code and pre-trained models.

\section{Challenge Results}
This year's challenge features two parallel tracks focusing on different aspects of image quality: Track 1 emphasizes restoration fidelity while Track 2 prioritizes perceptual quality. Tab.~\ref{tab:main_results} presents comprehensive rankings and performance metrics for all participants.

\noindent\textbf{Track 1 (Restoration Quality).} SamsungAICamera leads the restoration track with an impressive 33.73 dB PSNR, setting a new benchmark for the challenge. The competition remains highly competitive at the top, with I2WM\&JNU closely following at 33.45 dB and SR-Strugglers securing third position at 31.98 dB. A remarkable achievement this year is that three solutions exceed 33.00 dB, while an overwhelming majority of twenty-three teams deliver results beyond the 30.00 dB mark, demonstrating the maturity of modern super-resolution techniques.

\noindent\textbf{Track 2 (Perception Quality).} In the perceptual quality track, SamsungAICamera demonstrates versatility by claiming the top spot with a score of 4.7853, successfully balancing both objective metrics and subjective quality. VEPG follows closely with 4.7666, while HONORAICamera secures third place at 4.4787. The overall performance distribution shows strong results, with seven solutions scoring above 4.0 and fourteen surpassing 3.6, reflecting significant community-wide progress in perception.

Detailed evaluation methodologies for both tracks are elaborated in Sec.~\ref{sec:track}. In Sec.~\ref{sec:teams}, we present team contributions following the ranking order shown in Tab.~\ref{tab:main_results}, with particular emphasis on the top 5 performers. Space constraints necessitate moving descriptions of remaining participants to Sec.~A of the supplementary materials, while complete team membership details are available in Sec.~B.

\vspace{2.mm}
\subsection{Architectures and main ideas}
Throughout the challenge, participants introduced a diverse set of techniques to improve both restoration fidelity and perceptual quality. Based on the submitted methods, several representative technical trends can be summarized as:

\begin{enumerate}
    \item \textbf{Pre-trained Transformer-based restoration backbones remain the dominant choice.}
    Transformer-based image restoration models continue to serve as the foundation of many competitive solutions. In particular, HAT, SwinIR, HMANet, and PFT-SR were widely adopted due to their strong capability in modeling both local structures and long-range dependencies. Rather than designing entirely new backbones, many teams built upon publicly available pre-trained models and adapted them to the challenge setting through lightweight fine-tuning or partial parameter updating. This trend suggests that strong pre-trained restoration priors remain highly effective for $\times4$ super-resolution.

    \vspace{2.mm}
    \item \textbf{Inference-time optimization plays an increasingly important role in competitive performance.}
    Beyond network design, several teams improved results through carefully designed inference strategies. Representative techniques include geometric self-ensemble~\cite{timofte2016seven}, tiled inference with overlap, Gaussian-weighted stitching, dynamic reflection padding, and even weight interpolation between pre-trained and fine-tuned checkpoints. These methods improve robustness on high-resolution test images, mitigate border inconsistencies, and better adapt foundation restoration models to the target benchmark without changing the underlying architecture.

    \vspace{2.mm}
    \item \textbf{Two-stage pipelines that separate fidelity restoration and perceptual enhancement have become a prominent paradigm.}
    A clear trend in this year's challenge is the use of cascaded frameworks that first reconstruct a faithful intermediate image and then enhance perceptual realism in a second stage. In these pipelines, the first stage is typically a strong deterministic restoration model such as HAT, while the second stage relies on a generative prior to synthesize realistic details. This design explicitly addresses the perception--distortion trade-off by assigning fidelity and realism to different modules, leading to improved perceptual quality while preserving structure from the low-resolution input.

    \vspace{3.mm}
    \item \textbf{Generative priors based on diffusion and rectified flow models are becoming a key ingredient for perceptual SR.}
    Several teams incorporated large-scale generative backbones, including diffusion models and rectified flow transformers, into the super-resolution pipeline. Compared with conventional restoration-only models, these approaches are better at hallucinating visually pleasing textures and recovering natural-looking details. At the same time, participants also proposed task-specific adaptations such as LoRA tuning, conditional token injection, structure-aware guidance, and low-resolution image concatenation, in order to improve fidelity and reduce the structural inconsistency often introduced by generative models.

    \vspace{3.mm}
    \item \textbf{More explicit conditioning mechanisms are used to inject degradation, semantic, and structural information.}
    Recent methods are no longer limited to feeding only the low-resolution image into the network. Instead, some teams extracted degradation-aware descriptors, structure maps, or semantic guidance and injected them into the restoration model through cross-attention or conditional branches. Such explicit conditioning helps the network distinguish among different degradation patterns and preserve important geometric structures, especially in diffusion-based restoration pipelines.

    \vspace{3.mm}
    \item \textbf{Detail-aware objectives and residual refinement modules are effective for improving local reconstruction quality.}
    In addition to standard reconstruction losses, participants introduced losses and modules specifically designed to emphasize difficult high-frequency regions. Examples include focal frequency supervision, perceptual losses, gradient/detail-aware weighted losses, and residual correction branches that refine the base prediction. These designs demonstrate that, even when large pre-trained models provide strong global priors, dedicated local-detail enhancement mechanisms remain essential for recovering fine structures.

    \vspace{2.mm}
    \item \textbf{Training strategies increasingly emphasize staged adaptation and efficient fine-tuning.}
    Instead of full retraining, many teams adopted more economical optimization strategies such as partial fine-tuning, staged training, sequential LoRA adaptation, or restoration-first / perception-second optimization. This reflects a broader shift toward leveraging strong pre-trained foundations while using lightweight adaptation to target the specific evaluation objective of the challenge, whether it is PSNR-oriented restoration or perceptual enhancement.
\end{enumerate}

\subsection{Participants}
This year, the image SR challenge attracted \textbf{194} registered participants, with \textbf{31} teams submitting valid entries. Compared with previous editions, the 2026 challenge further demonstrated the growing interest in high-quality image super-resolution, especially in combining strong restoration backbones with modern generative priors. The submitted methods cover a broad spectrum, ranging from pure Transformer-based restoration models to diffusion-based and rectified-flow-based perceptual enhancement pipelines, reflecting the rapid evolution of the field.

\subsection{Fairness}
A set of rules has been established to ensure the fairness of the competition. \textbf{(1)} The use of benchmark test HR images for training is strictly prohibited, although the corresponding LR inputs may be available for evaluation purposes. \textbf{(2)} Participants are allowed to train with publicly available external datasets, provided that these datasets are properly declared and are consistent with the challenge rules. \textbf{(3)} Standard data augmentation techniques during training and inference, such as random flipping, rotation, self-ensemble, and tiled testing, are considered fair practice. \textbf{(4)} The use of publicly available pre-trained models and foundation backbones is allowed, as long as no forbidden test annotations or target-domain ground truth data are involved in training.

\subsection{Conclusions}
The insights gained from analyzing the results of the NTIRE 2026 image super-resolution (SR) challenge:
\begin{enumerate}
    \item Pre-trained Transformer-based restoration models remain the strongest foundation for high-fidelity super-resolution. Methods based on HAT, SwinIR, HMANet, and related architectures continue to provide competitive performance, especially when combined with careful fine-tuning and inference-time optimization.

    \item Two-stage pipelines that decouple fidelity restoration from perceptual detail generation have emerged as a particularly effective solution for perceptual SR. This design enables participants to preserve structural consistency in the first stage while exploiting strong generative priors in the second stage.

    \item Generative backbones, including diffusion and rectified-flow models, are playing an increasingly important role in perceptual super-resolution. Their effectiveness is further improved by task-specific adaptation techniques such as LoRA tuning, conditional guidance, and structure-aware generation.

    \item Explicit conditioning on degradation, structure, and semantic information helps modern SR models better handle diverse degradations and preserve important content, especially in conditional generative frameworks.

    \item Fine-grained detail enhancement remains a crucial factor. Frequency-aware supervision, gradient/detail-weighted losses, and residual refinement modules are effective complements to large pre-trained models, leading to improved recovery of textures and local structures.

    \item Efficient adaptation strategies, such as staged training, partial fine-tuning, and inference-time enhancement, have become increasingly important in practice. These strategies allow teams to fully exploit large pre-trained models while maintaining stability and achieving strong performance under challenge constraints.
\end{enumerate}

\section{Challenge Methods and Teams}
\label{sec:teams}
\subsection{SamsungAICamera}
\begin{figure}[t]
  \centering
  \includegraphics[width=0.85\linewidth]{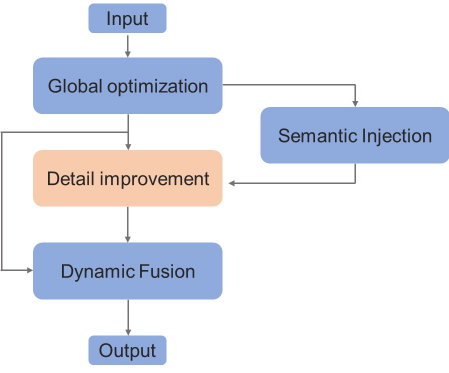}
  \caption{\textbf{Team SamsungAICamera}}
  \label{fig:team09-samsung}
  \vspace{-4mm}
\end{figure}

\noindent\textbf{Description.}
The SamsungAICamera team proposes a cascaded hybrid super-resolution framework that aims to balance global structural reconstruction and local texture enhancement, illustrated in Fig.~\ref{fig:team09-samsung}. The model first extracts shallow features from the low-resolution input and feeds them into a Global Optimization Module (GOM), which is built upon the HAT architecture~\cite{chen2023hat}. As the primary backbone, the GOM is responsible for capturing long-range spatial dependencies and reconstructing the fundamental geometric structure of the target high-resolution image.

To further improve local detail restoration, the team introduces a Detail Enhancement Module (DEM) based on NAFNet~\cite{chen2022simple}. Instead of simply forwarding the final output of the GOM to the second stage, the method extracts deep intermediate features from the later GOM blocks and injects them into the DEM through a Semantic Injection Module (SIM). In this way, semantically rich global representations are used to guide the restoration of high-frequency details and fine textures in the local enhancement stage.

Finally, the outputs of the two branches are adaptively combined by a Dynamic Fusion Module (DFM). Specifically, the DFM introduces two independent spatial weight maps to aggregate the outputs of the GOM and DEM, enabling the network to selectively preserve structural information and enhance local textures. This design allows the model to leverage the complementary strengths of transformer-based global modeling and CNN-based local refinement within a unified framework.

\noindent\textbf{Implementation Details.}
The team trained the model using three datasets: DIV2K, LSDIR, and a self-collected dataset containing 2 million images. The overall training pipeline consists of two stages. In the first stage, the entire network is pre-trained on the 2M-image custom dataset with LSDIR using an initial learning rate of $1\times10^{-4}$ for approximately 360 hours. In the second stage, the model is fine-tuned on DIV2K with an initial learning rate of $1\times10^{-5}$ for around 120 hours, which further improves detail restoration.

During training, the team employs data augmentation strategies including channel shuffle and mixing. They also adopt a progressive learning strategy by gradually increasing the patch size from $320\times320$ to $448\times448$, and finally to $768\times768$, which consistently improves performance.

For optimization, the model is first trained by alternating among $L_1$ loss, $L_2$ loss, and Stationary Wavelet Transform (SWT) loss~\cite{korkmaz2024training}. According to the team, incorporating SWT loss helps the optimization escape local optima. To further improve perceptual quality and better align the model with the challenge evaluation criteria, they subsequently introduce a set of IQA-oriented loss functions, including LPIPS, DISTS, CLIP-IQA, MANIQA, MUSIQ, and NIQE. Since these losses have different numerical scales and gradient magnitudes, the team performs empirical balancing and grid search to determine suitable loss weights. In addition, because the NIQE implementation in the \texttt{pyiqa} library is unstable during backpropagation, they replace it with a differentiable approximation. All models are trained on an NVIDIA A100 80GB GPU.

\subsection{I2WM\&JNU}
\begin{figure}[t]
    \centering
    \includegraphics[width=1\linewidth]{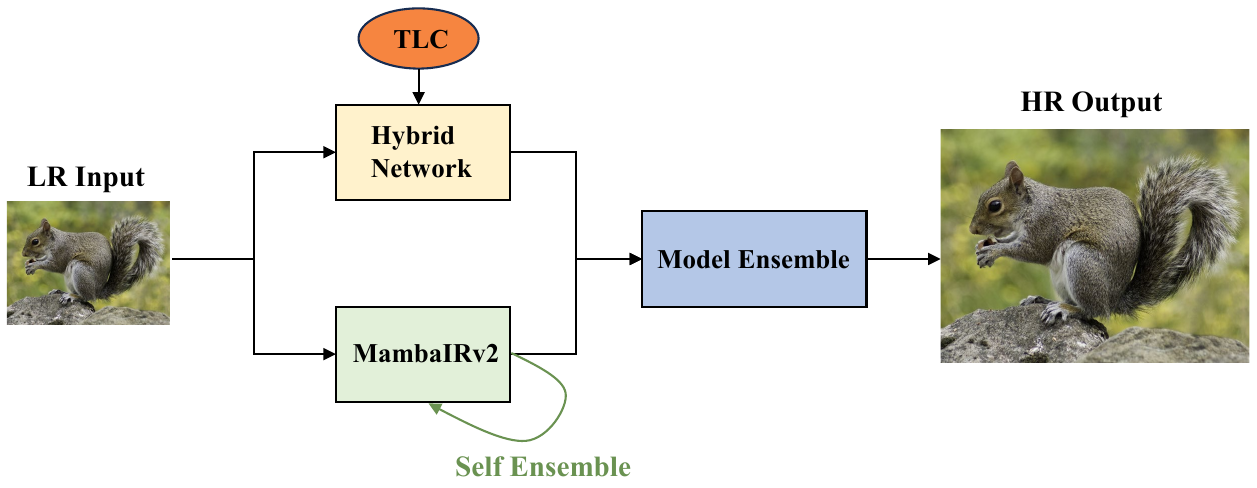}
    \caption{\textbf{Team I2WM\&JNU}}
    \label{fig:team02-i2wmjnu}
    \vspace{-6mm}
\end{figure}

\noindent\textbf{Description.}
The proposed solution is illustrated in Fig.~\ref{fig:team02-i2wmjnu}. Instead of training a new super-resolution network from scratch, the I2WM\&JNU team adopts an inference-only strategy that leverages two powerful pre-trained models, namely the Hybrid Network (SRC-B) and MambaIRv2. The key idea is to exploit the complementary strengths of these two architectures through inference-time enhancement and model ensembling, thereby achieving competitive super-resolution performance without additional cost.

Specifically, the team applies Test-time Local Converter (TLC) to the Hybrid Network in order to improve local reconstruction quality, while a self-ensemble strategy~\cite{timofte2016seven} is employed for MambaIRv2 to enhance robustness and prediction quality. The resulting outputs from the two enhanced models are then combined using a weighted model ensemble to produce the final super-resolved image. By directly leveraging existing strong priors embedded in the pre-trained models, this solution provides an alternative to computationally expensive retraining pipelines.

\noindent\textbf{Implementation Details.}
The method uses the officially released pre-trained models of Hybrid Network (SRC-B) and MambaIRv2. The Hybrid Network model is pre-trained on DIV2K and LSDIR, while MambaIRv2 is pre-trained on DIV2K and Flickr2K. No additional training or fine-tuning is performed for this challenge submission.

During inference, TLC is applied to the Hybrid Network to refine local details, and self-ensemble~\cite{timofte2016seven} is applied to MambaIRv2 to improve prediction stability. Their weighted average yields the final output. This design keeps the overall pipeline simple and training-free, while still benefiting from the complementary characteristics of two state-of-the-art super-resolution models.

\subsection{VEPG}
\begin{figure}[t]
    \centering
    \includegraphics[width=\linewidth]{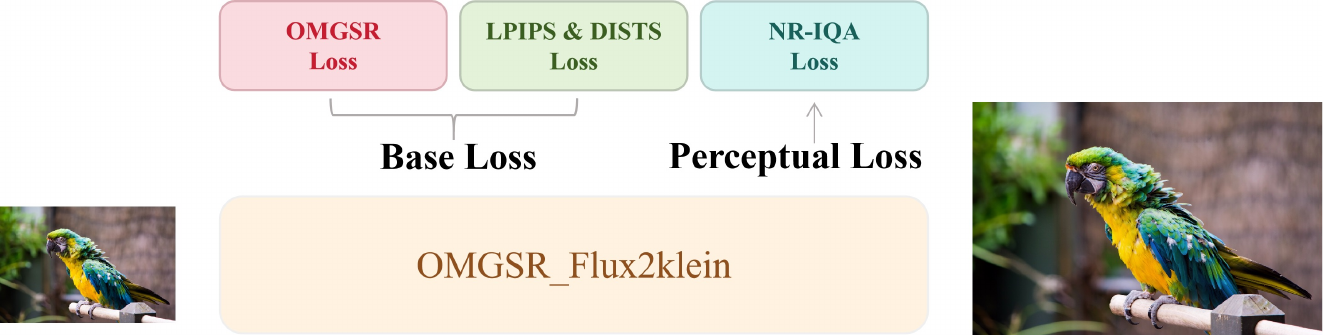}
    \caption{\textbf{Team VEPG}}
    \label{fig:team05-vepg}
    \vspace{-4.mm}
\end{figure}

\noindent\textbf{Description.}
The proposed solution is illustrated in Fig.~\ref{fig:team05-vepg}. The VEPG team builds its method upon OMGSR~\cite{wu2025omgsrneedmidtimestepguidance}, a one-step diffusion-based super-resolution framework that injects low-quality image information by adjusting the diffusion timestep associated with the input. To further enhance perceptual quality, the team adopts FLUX.2-klein-base (4B)~\cite{flux2} as the backbone model.

To improve the perceptual relevance of the optimization objective, the team replaces the original DINOv3-based DISTS loss in OMGSR with LPIPS~\cite{zhang2018perceptual} and DISTS~\cite{dists_loss}, and combines them with the original restoration-oriented objectives. The resulting base loss is formulated as
\begin{equation}
\begin{aligned}
\mathcal{L}_{\mathrm{base}}
=&\;
\lambda_{1} \mathcal{L}_{\mathrm{LRR}}
+
\lambda_{2} \mathcal{L}_{\mathrm{L1}}
+
\lambda_{3} \mathcal{L}_{\mathrm{GAN}} \\
&+
\lambda_{4} \mathcal{L}_{\mathrm{LPIPS}}
+
\lambda_{5} \mathcal{L}_{\mathrm{DISTS}},
\end{aligned}
\end{equation}
where $\mathcal{L}_{\mathrm{LRR}}$ aligns the low-quality input at the specified timestep with the noised high-quality target, $\mathcal{L}_{\mathrm{L1}}$ preserves pixel-level fidelity, and $\mathcal{L}_{\mathrm{GAN}}$ improves visual realism.

Since OMGSR is a one-step diffusion framework built on a pretrained model, the team further introduces no-reference image quality assessment (NR-IQA) losses~\cite{ke2021musiq,wang2022exploring,yang2022maniqa} to enhance perceptual quality. Following this design, the perceptual supervision term is defined as
\begin{equation}
\begin{aligned}
\mathcal{L}_{\mathrm{NR\mbox{-}IQA}}
=&\;
\lambda_{6} \mathcal{L}_{\mathrm{MUSIQ}}
+
\lambda_{7} \mathcal{L}_{\mathrm{CLIPIQA}} \\
&+
\lambda_{8} \mathcal{L}_{\mathrm{MANIQA}},
\end{aligned}
\end{equation}
where all three IQA terms are differentiable and implemented mainly based on the \texttt{pyiqa} library. For metrics in which larger scores indicate better quality, the team normalizes the scores to $[0,1]$ and converts them into minimization objectives using $1-s$. The overall training objective is
\begin{equation}
\mathcal{L}_{\mathrm{total}}
=
\mathcal{L}_{\mathrm{base}}
+
\mathcal{L}_{\mathrm{NR\mbox{-}IQA}}.
\end{equation}

\noindent\textbf{Implementation Details.}
The model is trained using the AdamW optimizer with $\beta_1=0.9$, $\beta_2=0.999$, and a learning rate of $2\times10^{-5}$. The ground-truth images are cropped into patches of size $512\times512$. Training is conducted for 8{,}000 iterations with a batch size of 4. The loss weights are set to $\lambda_1=3$, $\lambda_2=0.5$, $\lambda_3=0.5$, $\lambda_4=5$, $\lambda_5=1$, $\lambda_6=0.8$, $\lambda_7=1$, and $\lambda_8=0.8$. Both the VAE rank and the transformer rank are set to 64.

Following FaithDiff~\cite{chen2025faithdiff}, the team uses the same training datasets, including DIV2K, LSDIR, and additional high-quality image datasets. This training configuration combines the restoration capability of one-step diffusion models with stronger perceptual supervision from both full-reference and no-reference image-quality objectives.

\subsection{SR-Strugglers}
\begin{figure}[t]
\centering
\includegraphics[width=\linewidth]{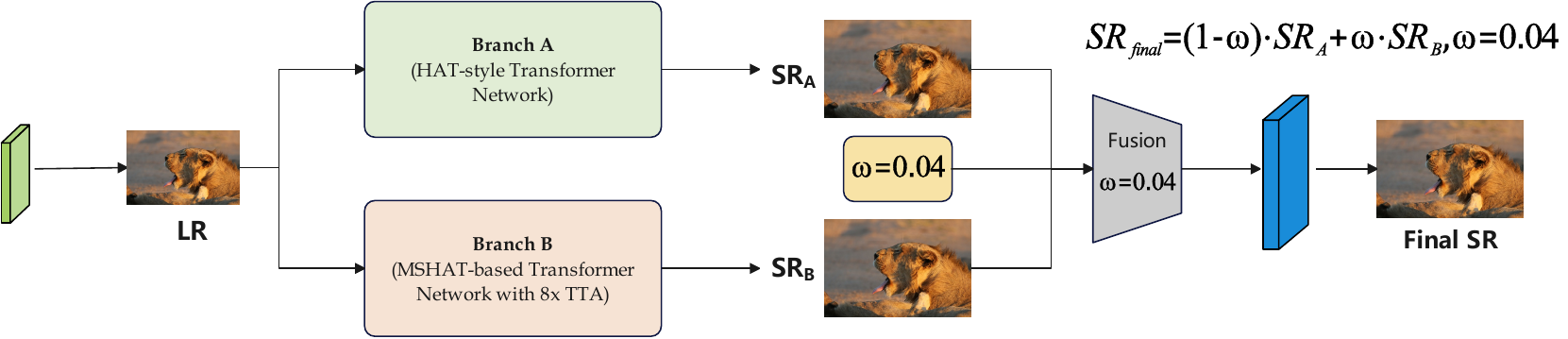}
\vspace{-1.mm}
\caption{\textbf{Team SR-Strugglers}}
\label{fig:team30-srstrugglers}
\vspace{-5.mm}
\end{figure}

\noindent\textbf{Description.}
The proposed solution is illustrated in Fig.~\ref{fig:team30-srstrugglers}. The SR-Strugglers team proposes FusionHero, a two-branch fusion framework SR. Given a low-resolution input image, the method processes it in parallel using two independent pretrained transformer-based super-resolution models. Branch~A is a HAT-style hybrid attention transformer that produces a first super-resolved estimate, while Branch~B is an MSHAT-based transformer whose outputs are further enhanced using $8\times$ test-time augmentation based on flips and transpose-based transforms.

The final prediction is obtained through a fixed global pixel-wise fusion of the two branch outputs. Specifically, the team combines the HAT-style output and the MSHAT-based output using a weighted average, where the fusion weight is set to $w=0.04$ according to validation-set performance. According to the team, the main challenge is not designing new model components, but rather identifying a pair of complementary pretrained models and determining an effective fusion weight. This design aims to combine the stable restoration capability of the main branch with the additional detail enhancement brought by the second branch and its test-time augmentation.

\noindent\textbf{Implementation Details.}
The method is inference-only and does not involve any additional training. Both branches use publicly available pretrained $\times4$ SR checkpoints. The only tuned hyperparameter is the fusion weight $w$, which is selected on the validation set to maximize PSNR.

At inference time, each low-resolution input image is first processed once by the HAT-style branch. In parallel, the MSHAT-based branch is applied under $8\times$ geometric test-time augmentation, where the input is transformed using flips and transpose-based variants, and the transformed predictions are averaged after inverse transformation. The final super-resolved result is then computed as the weighted average of the outputs from the two branches. The team reports that the small contribution from Branch~B improves detail enhancement while limiting its impact on the dominant restoration result from Branch~A.

The team does not use any extra training data, since no training is performed. The validation set is used solely for selecting the fusion weight, and testing is conducted directly on the challenge test images. The total model complexity is approximately the sum of the two transformer branches, with each branch containing on the order of tens of millions of parameters. Runtime depends on hardware, and the use of $8\times$ test-time augmentation increases the computational cost of Branch~B accordingly.

\subsection{HONORAICamera}
\begin{figure}[t]
\centering
\includegraphics[width=\linewidth]{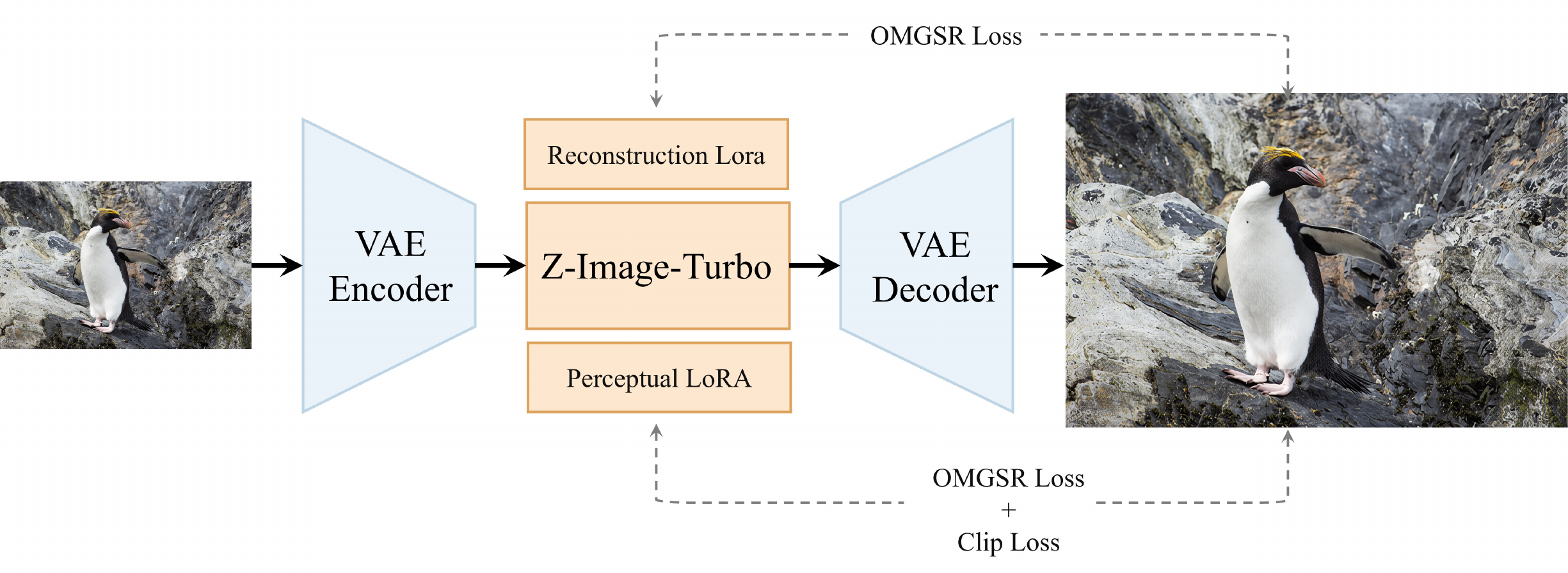}
\vspace{-1.mm}
\caption{\textbf{Team HONORAICamera}}
\label{fig:team14-HONORAICamera}
\vspace{-5mm}
\end{figure}

\noindent\textbf{Description.}
The HONORAICamera team proposes a diffusion-based generative prior framework for real-world image super-resolution, as shown in Fig.~\ref{fig:team14-HONORAICamera}. Their method is built upon Z-Image-Turbo~\cite{zit}, which provides a latent generative backbone for high-quality image synthesis. To adapt this pretrained model to the super-resolution task, the team follows the training strategy of OMGSR~\cite{wu2025omgsrneedmidtimestepguidance}, aiming to reconstruct high-resolution details from low-resolution inputs while preserving structural consistency.

Specifically, the method adopts the architecture of Z-Image-Turbo and fixes both the training and inference timestep to 121. This single-step setting is designed to balance reconstruction quality and computational efficiency, while effectively generating high-frequency details.

\noindent\textbf{Implementation Details.}
The training pipeline consists of two stages. In the first stage, the team focuses on transferring the pretrained generative prior to the super-resolution task. They train the model on LSDIR, DIV2K\_train, Flickr2K\_train, and the first 10,000 images from FFHQ. The training resolution is set to $1024\times1024$, with a total batch size of 128, and the model is trained for 6,000 steps. The objective of this stage is to align the latent generative space with the degraded low- and high-resolution image pairs.

In the second stage, the team further optimizes perceptual quality and no-reference image quality metrics. For this purpose, they use DIV2K\_train, Flickr2K\_train, and a small collection of custom high-quality DSLR images captured by the team. Following the standard OMGSR framework, they employ MSE, Dv3D, GAN, and LRR losses for training. In addition, to further improve the CLIP-IQA score in the challenge setting, they introduce an extra CLIP loss to supervise semantic consistency. According to the team, while this additional loss is effective for improving no-reference metrics, it may also introduce pseudo-textures and high-frequency artifacts. In practical applications, removing the CLIP loss leads to more natural textures and better visual quality, revealing a trade-off between metric-oriented optimization and perceptual fidelity.

For both training stages, the Real-ESRGAN pipeline synthesizes low-resolution inputs, comprising various blur kernels, Gaussian and Poisson noise, and JPEG compression artifacts to simulate realistic degradations.

\subsection{IK-LAB}
\begin{figure}[t]
    \centering
    \includegraphics[width=0.49\textwidth]{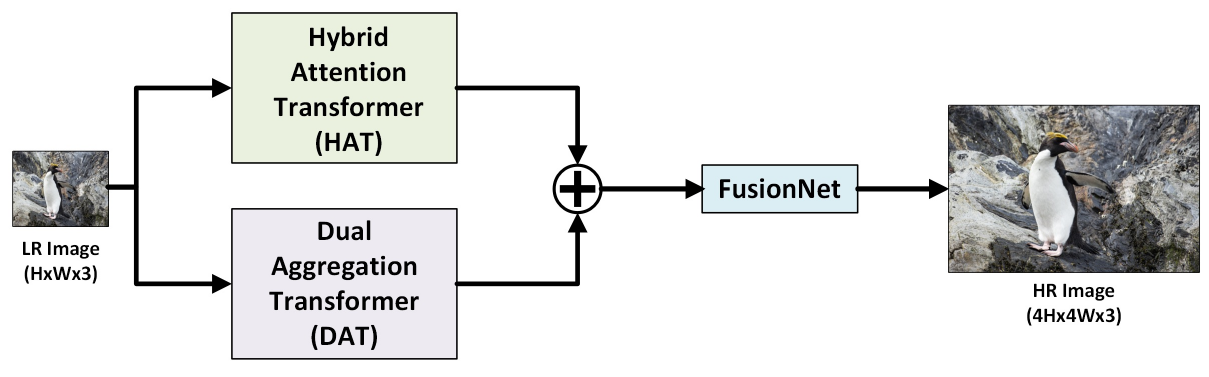}
    \vspace{-4.mm}
    \caption{\textbf{Team IK-LAB}}
    \label{fig:team25-iklab-fig1}
    \vspace{-6.mm}
\end{figure}

\noindent\textbf{Description.}
The proposed method is illustrated in Fig.~\ref{fig:team25-iklab-fig1}. The IK-LAB team combines two frozen pretrained super-resolution backbones, HAT-IQCMix~\cite{chen2023hat} and DAT~\cite{chen2023dat}, through a lightweight trainable fusion network, denoted as FusionNet. The key idea is that the two transformer-based backbones employ different attention mechanisms and therefore produce complementary reconstruction errors, which can be exploited through adaptive fusion. Only the fusion module is trained, while both pretrained backbones remain fixed, in order to preserve their original restoration capability and reduce the risk of overfitting.

HAT-IQCMix is based on the Hybrid Attention Transformer and incorporates image-quality-adaptive conditional mixing. It contains 12 Residual Hybrid Attention Groups (RHAGs) with an embedding dimension of 180. The first 10 RHAGs form a shared trunk, while the final 2 RHAGs are replicated into 10 branch-specific tails corresponding to five image quality attributes and two threshold branches. During inference, five quality metrics, including brightness, contrast, sharpness, noise, and saturation, are computed from the input image, and the selected branch outputs are averaged. Contrastingly, DAT alternates between spatial and channel attention within each DAT block. It consists of 6 residual groups with 6 DAT blocks each, using an embedding dimension of 180 and 6 attention heads. Both backbones use PixelShuffle for $\times4$ upsampling.

FusionNet takes the concatenated RGB outputs of the two backbones as a 6-channel input. It first encodes these features using two $3\times3$ convolution layers with channel dimensions $6\rightarrow64\rightarrow32$ and LeakyReLU activation, followed by squeeze-and-excitation channel attention. A prediction head then generates per-pixel softmax weights over the two backbone outputs. In addition, a residual refinement branch is introduced and scaled by a learnable factor. The final output is clamped to the valid image range. At inference time, the team further applies the $8\times$ geometric self-ensemble~\cite{timofte2016seven} based on four rotations and two flips, and averages the inverse-transformed results pixel-wise.

\noindent\textbf{Implementation Details.}
FusionNet is trained on the DIV2K training set with $\times4$ bicubic downsampling, using fixed NTIRE 2025 weights for both backbones. No external training data is used. During training, random $128\times128$ low-resolution patches, corresponding to $512\times512$ high-resolution patches, are cropped with random horizontal and vertical flips. Reflection padding is applied to align inputs to multiples of 16 for window-based attention, and the padding is removed after super-resolution.

The training objective consists of a Charbonnier loss, an FFT-domain loss, and a Sobel-gradient loss. Specifically, the total loss is defined as
\begin{equation}
    \mathcal{L}_{\text{total}} = \mathcal{L}_{\text{Charb}} + \lambda_{\text{FFT}} \cdot \mathcal{L}_{\text{FFT}} + \lambda_{\text{Sobel}} \cdot \mathcal{L}_{\text{Sobel}},
\end{equation}
where the Charbonnier loss is used for robust pixel-level reconstruction, the FFT loss enforces consistency in the frequency domain, and the Sobel loss improves gradient and edge fidelity. The team notes that image-quality-assessment-oriented losses such as LPIPS, DISTS, CLIP-IQA, MANIQA, MUSIQ, and NIQE are not used in the current setting, which they identify as one reason for suboptimal performance on Track 2.

The model is optimized using Adam with $\beta_1=0.9$ and $\beta_2=0.999$. The initial learning rate is set to $1\times10^{-4}$ and annealed to $1\times10^{-6}$ using cosine scheduling over 100 epochs. The batch size is 16, and validation is conducted every 3 epochs on the DIV2K validation set, with the best checkpoint selected according to PSNR-Y. All experiments are performed on two NVIDIA RTX 4090 GPUs.

\section{Methods of the Remaining Teams}
The participating teams proposed diverse solutions and conducted extensive experimental studies throughout the competition. Owing to space constraints, a more comprehensive description is provided in Sec.~\textcolor{red}{A} of the supplementary materials, where the methods and implementation details of the remaining teams are presented in detail. The supplementary material is available at the project page\footnote{\url{https://github.com/zhengchen1999/NTIRE2026_ImageSR_x4/releases/download/v1/NTIRE2026_ImageSR_x4_Supplementary_Material.pdf}}. Although these teams are not covered in the main report, their contributions provide valuable insights into different technical designs.

\section*{Acknowledgments}
This work is partially supported by the Humboldt Foundation. We thank the NTIRE 2026 sponsors: OPPO, Kuaishou, and the University of Wurzburg (Computer Vision Lab). 
This work is supported by the National Natural Science Foundation of China (62501386, 625B2116, 625B1025), CCF-Tencent Rhino-Bird Open Research Fund. This work is also sponsored by Al Hundred Schools Program and is carried out using the Ascend AI technology stack.

{\small
\bibliographystyle{ieeenat_fullname}
\bibliography{main}
}

\end{document}